\documentclass[letterpaper, 10 pt, conference]{ieeeconf}  

\IEEEoverridecommandlockouts                              

\usepackage{caption}
\usepackage{graphicx}
\usepackage{float} 
\usepackage{subfigure}
\usepackage{amsmath,amssymb,amsfonts}
\usepackage{xcolor}
\usepackage{color}
\usepackage[switch]{lineno}
\usepackage{marvosym}
\usepackage{textcomp}
\usepackage{verbatim}
\usepackage{hyperref}
\usepackage{array}
\usepackage{booktabs,multirow,bm}
\usepackage{tabularx}
\usepackage{multirow}

\title{\LARGE \bf
Never too Prim to Swim: An LLM-Enhanced RL-based Adaptive S-Surface Controller for AUVs under Extreme Sea Conditions
}

\author{Guanwen Xie$^{1,*}$, Jingzehua Xu$^{1,*,}\textsuperscript{\Letter}$, Yimian Ding$^{1}$, Zhi Zhang$^{1}$, Shuai Zhang$^{2}$ and Yi Li$^{1,}\textsuperscript{\Letter}$
\thanks{*These authors contribute to this work equally.}
\thanks{\textsuperscript{\Letter}Corresponding authors.}
\thanks{$^{1}$G. Xie, J. Xu, Y. Ding, Z. Zhang and Y. Li are with Tsinghua Shenzhen International Graduate School, Tsinghua University, Shenzhen, 518055, China. E-mail: \{xgw24, xjzh23, dingym24, z-zhang23\}@mails.tsinghua.edu.cn, liyi@sz.tsinghua.edu.cn.}%
\thanks{$^{2}$S. Zhang is with Department of Data Science, New Jersey Institute of Technology, NJ 07102, USA. E-mail: sz457@njit.edu.}
}

\begin{document}

\maketitle
\thispagestyle{empty}
\pagestyle{empty}

\setcounter{footnote}{2}
\begin{abstract}

The adaptivity and maneuvering capabilities of Autonomous Underwater Vehicles (AUVs) have drawn significant attention in oceanic research, due to the unpredictable disturbances and strong coupling among the AUV's degrees of freedom. In this paper, we developed large language model (LLM)-enhanced reinforcement learning (RL)-based adaptive S-surface controller for AUVs. Specifically, LLMs are introduced for the joint optimization of controller parameters and reward functions in RL training. Using multi-modal and structured explicit task feedback, LLMs enable joint adjustments, balance multiple objectives, and enhance task-oriented performance and adaptability. In the proposed controller, the RL policy focuses on upper-level tasks, outputting task-oriented high-level commands that the S-surface controller then converts into control signals, ensuring cancellation of nonlinear effects and unpredictable external disturbances in extreme sea conditions. Under extreme sea conditions involving complex terrain, waves, and currents, the proposed controller demonstrates superior performance and adaptability in high-level tasks such as underwater target tracking and data collection, outperforming traditional PID and SMC controllers. \footnote{The accompanying videos, details about prompts and LLM responses, and source code are available at the website \href{https://360zmem.github.io/AUV-RSControl}{\textbf{https://360zmem.github.io/AUV-RSControl}} .}

\end{abstract}

\section{INTRODUCTION}

The adaptive control and maneuvering capabilities of Autonomous Underwater Vehicles (AUVs) have drawn significant attention in oceanic research due to their substantial potential in maritime applications, including underwater resource exploration \cite{1}, shipwreck search \cite{2}, and underwater structure maintenance \cite{3}. These capabilities contribute significantly to marine science and the economy \cite{4}, but require advanced control systems that provide task-adaptive and precise control of AUVs' position and attitude, particularly under extreme sea conditions \cite{5}. However, achieving precise maneuvering control of AUVs is challenging due to their highly nonlinear dynamics \cite{6}, time-varying hydrodynamics, strong six-degree-of-freedom coupling, and environmental uncertainties \cite{7}. During ocean navigation, AUVs encounter unpredictable external disturbances \cite{8}, requiring continuous high-precision trajectory tracking and obstacle avoidance during tasks such as coral reef ecosystem monitoring \cite{9}, which necessitates balancing multiple objectives \cite{11}.  Additionally, position uncertainties caused by extreme sea conditions \cite{10} require additional control compensation.

    \begin{figure}[!t]
        \centering
        \includegraphics[width=0.984\linewidth]{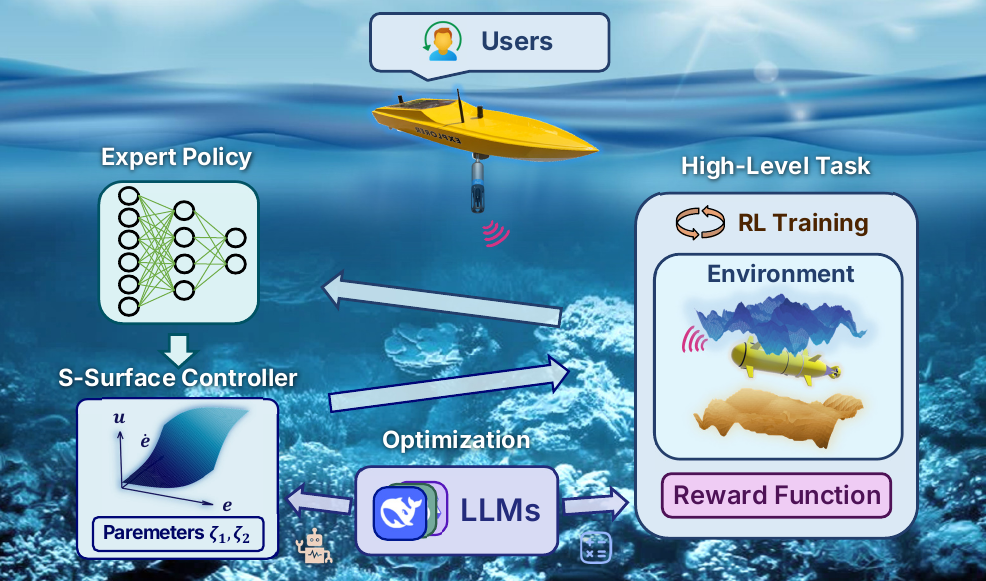}
        \caption{\small \textbf{Illustration of an AUV conducting underwater tasks using the proposed controller}. The proposed controller utilizes an RL-based S-surface controller to enable effective control. LLMs assist the controller by optimizing the reward functions for RL training and tuning the parameters of the S-surface controller.}
        \label{fig_1}
        \end{figure}

Researchers have developed various control methodologies for AUVs, including PID controllers, sliding mode control (SMC) \cite{12}, fuzzy control \cite{13}, and model predictive control (MPC) \cite{14}. While these methods demonstrate advantages in most scenarios, they exhibit limited adaptability in extreme conditions. Specifically, PID controllers require time-consuming parameter tuning for complex environments \cite{15}. Fuzzy controllers provide good stability but are limited by the complexity of defining membership functions, inference methods, and fuzzy rules \cite{13}. MPC predicts future behavior for optimized control but heavily relies on real-time computation and accurate system models, reducing its robustness in extreme conditions \cite{14}.

The S-Surface controller has shown promise in handling uncertainties and nonlinearities, which leverages a sigmoid plane-like surface to control AUVs’ dynamic systems towards desired states \cite{16}. However, it lacks the flexibility to adaptively adjust parameters and control strategies to handle the strong coupling between degrees of freedom \cite{17}. The emergence of Reinforcement Learning (RL) has somehow addressed these issues. By training robots to learn adaptive control strategies through environmental interactions, RL has shown promising results in various applications including drone control \cite{18}, legged robot navigation \cite{19}, and other autonomous systems \cite{20}. Although RL faces challenges like reward function design, its strong learning ability enables AUVs to develop expert-level control strategies that autonomously map high-level 6-DoF commands to end-to-end control signals, including thruster commands \cite{7}. Also, with the assistance of the Large Language Model (LLM), AUVs can adaptively adjust controller’s parameters while optimizing RL reward functions \cite{22}, enhancing AUVs' ability to balance multi-objective optimization and improve task-oriented control and maneuvering capabilities in extreme marine environments \cite{7, 14}.

Based on above analysis, we develop an LLM-enhanced RL-based adaptive S-surface controller for AUVs to effectively execute high-level tasks in extreme sea conditions. The contributions of this paper mainly include three parts: 

\begin{itemize}
\item 
We develop a novel AUV controller that employs RL to train an expert-level control strategy for high-level task execution and control command generation, while the S-surface controller produces control signals, ensuring cancellation of nonlinear effects and external disturbances under extreme sea conditions. 
\item
We utilize LLMs for joint optimization of RL reward function and controller parameters, utilizing multimodal task execution logs and combining contextual information such as environmental descriptions to enhance the final task performance and adaptability.
\item
The proposed controller demonstrates superior robustness and flexibility compared to conventional PID and SMC controllers in challenging marine conditions characterized by waves, currents, and complex terrain. It exhibits exceptional performance in advanced 3D tasks, including underwater target tracking and data collection tasks.
\end{itemize}

\section{RELATED WORK}

\subsection{S-Surface Controller for AUV Control}

S-Surface controller and its variants leverage the principles of smooth surfaces and dynamic control, significantly enhancing AUV maneuverability and environmental disturbance responsiveness. Li \textit{et al}. \cite{6} implemented the controller on MOOS-IvP, demonstrating robust lake test results despite buoyancy variations. Lakhekar \textit{et al}. \cite{8} combined disturbance-observer-based control with fuzzy-adaptive S-Surface control for trajectory tracking, effectively compensating for disturbances without prior knowledge of uncertainty bounds. Jiang \textit{et al}. \cite{26} enhanced the S-Surface controller with a sliding mode variable structure to handle static load and high-speed motion, with stability confirmed by Lyapunov analysis.

\subsection{Reinforcement Learning for Control}

RL methods demonstrate promising results in controlling complex robotic systems, especially in challenging environments. Meger \textit{et al}. \cite{28} employed an RL-based approach to control a flipper-based underwater vehicle, using a Gaussian process model to predict state distributions. Hadi \textit{et al}. \cite{29} investigated RL for learning 2-DoF control (yaw, speed) in a simulator. Lu \textit{et al}. \cite{30} applied domain randomization to enhance RL-based control for a 4-DoF AUV. Notably, RL is often applicable to various settings without requiring in-situ tuning \cite{27}.

\subsection{Large Language Model for Multi-Objective Optimization}

LLMs excel in multi-objective optimization, serving as high-level semantic planners for robotic tasks \cite{31}, learning complex manipulation tasks, and generating structured outputs for sequential decision-making \cite{32}. Ma \textit{et al}. \cite{33} showed that LLM-generated rewards outperformed human-engineered ones across various robotic tasks. Xie \textit{et al}. \cite{34} utilized LLMs for creating interpretable, dense reward codes, enabling iterative refinement for multi-objective tasks with human feedback. Zarzà \textit{et al}. \cite{35} used GPT-3.5-turbo for instantaneous PID system updates, highlighting its network control potential. Guo \textit{et al}. \cite{36} leveraged LLMs to encode expert knowledge, emulating human-like gradual tuning of controller parameters to meet stability requirements.

\section{CONTROLLER DESIGN}

In this section, we detail our proposed controller, describing its overall design architecture and explaining the workflow and principles of its three main modules. 

\subsection{Structure of the Proposed Controller}

Fig. 2 illustrates the overall design of our controller. To fully leverage the advantages of the LLM-enhanced RL-based S-Surface controller, while achieving simulation and perception of extreme marine conditions to evaluate the disturbance rejection performance, we decompose the proposed framework into three core modules. Specifically, the \textbf{RL-based S-Surface Controller Module} employs RL policies focusing on high-level task decision-making, and the S-Surface controller utilized to achieve precise 6-DoF control. The \textbf{LLM-enhanced Iterative Joint Optimization Module} performs joint optimization of the RL reward function and controller parameters guided by domain-specific guidelines. It systematically analyzes environmental summaries, numerical computations, and multi-modal task feedback to enhance adaptation to dynamic marine environments. The \textbf{Simulation and Environment-Aware Module} executes physical ocean modeling with 6-DoF control dynamics for extreme scenario simulation, and fuses multisource sensor data for active disturbance mitigation.

    \begin{figure*}[!t]
        \centering
        \includegraphics[width=0.909\linewidth]{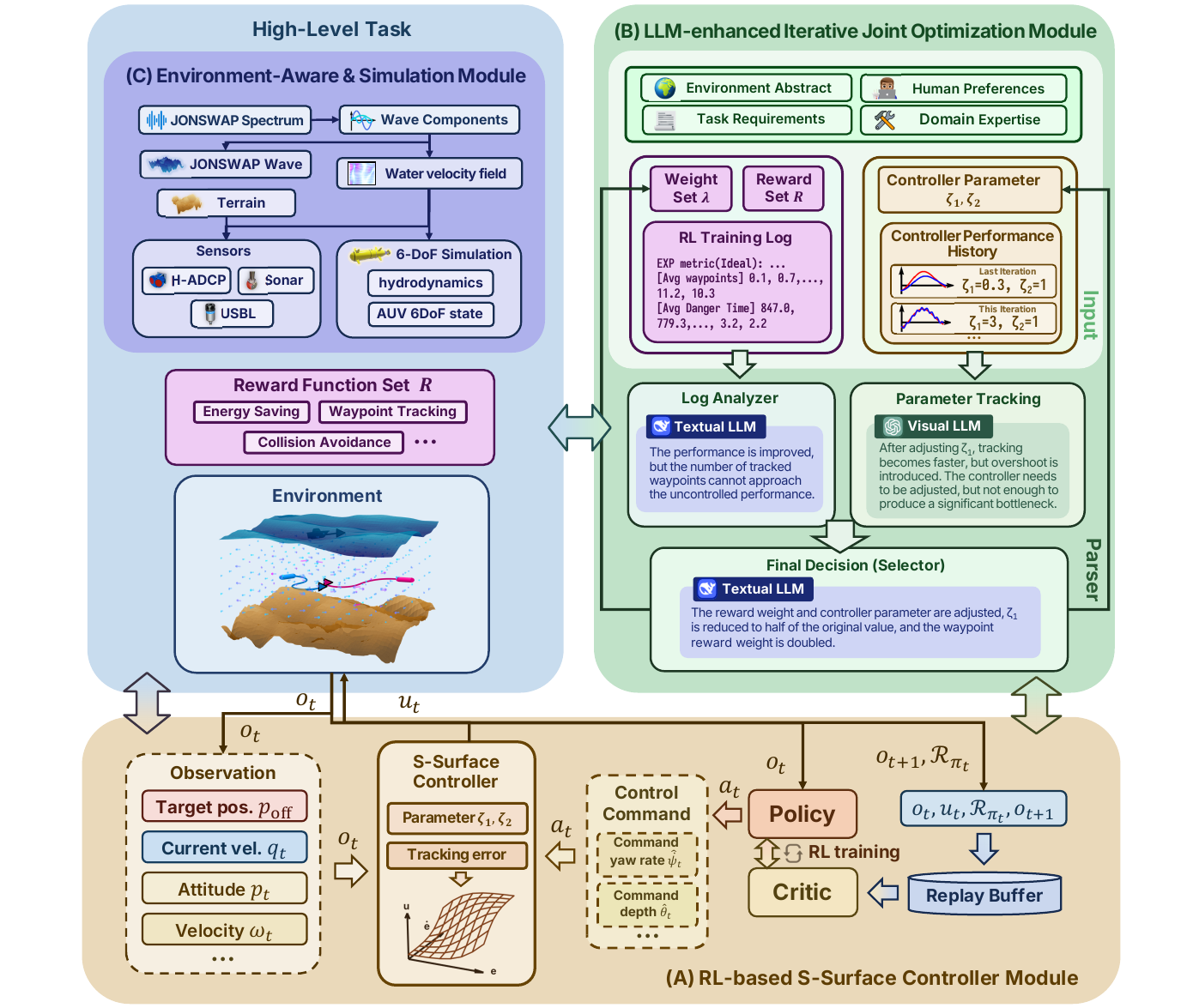}
        \caption{\small \textbf{The overall framework of our proposed controller}, which comprises three modules: (A) RL-based S-Surface Controller Module. (B) LLM-Enhanced Iterative Joint Optimization Module. (C) Environment-Aware and Simulation Module.}
        \label{fig_2}
    \end{figure*}

\subsection{RL-based S-Surface Controller Module} 
Through RL training, we aim to learn expert-level control policies optimized for end-to-end performance in control-constrained systems. The policies should demonstrate disturbance rejection capabilities and generate optimal reference signals for subordinate S-Surface controllers to enable AUVs to accomplish high-level tasks.

\textit{\textbf{Markov decision process modeling}}: We define the RL training process using a Markov decision process (MDP) with control-affine dynamics, represented as the tuple $\mathcal{M} \triangleq \left(\right. \mathcal{X} ,\mathcal{A}, \mathcal{U}, C, f , g , d , \mathcal{R}_{\pi} , \gamma \left.\right)$. Here, $\mathcal{X} \subseteq \mathbb{R}^{n}$ represents the state space, $\mathcal{A} \subseteq \mathbb{R}^{a}$ denotes the action space, while $\mathcal{U} \subseteq \mathbb{R}^{m}$ denotes the control signal space. The state transitions in the MDP follow the control-affine system:
\begin{equation}
x_{t + 1} = f \left(x_{t}\right) + g \left(x_{t}\right) C(a_t) + d \left(x_{t}\right),
\end{equation}
where $x_{t} \in \mathcal{X}$ represents the state at time step $t$. The high-level action signal is sampled from the distribution $\pi (a_{t} | x_{t})$ according to a RL control policy $\pi$, and the control signal $u_t=C(x_t,a_t)$ is generated by the controller $C:\mathcal{X} \times \mathcal{A}\to \mathcal{U}$. The functions $f:\mathbb{R}^{n} \rightarrow \mathbb{R}^{n}$ and $g : \mathbb{R}^{n} \rightarrow \mathbb{R}^{n \times m}$ characterize the known nominal model of the system. Additionally, $d:\mathbb{R}^{n} \rightarrow \mathbb{R}^{n}$ represents the unknown model component, such as environmental disturbances like ocean waves, which is continuous with respect to the state. The variable $\mathcal{R}_{\pi}$ denotes the reward functions, and $\gamma \in [0,1]$ is the discount factor.

According to Eq. (1), the transition probability is represented as $P ( x_{t + 1} \left| x_{t} , a_{t}\right)$. The closed-loop transition probability under policy $\pi$ is expressed as $P_{\pi} \left(x_{t + 1}|x_{t}\right) \triangleq \int_{\mathcal{U}}^{} \pi \left(\right. a_{t} \left|\right. x_{t} \left.\right) P \left(\right. x_{t + 1} \left|\right. x_{t} , a_{t} \left.\right) d a_{t}$. Furthermore, the closed-loop state distribution at time step 
$t$ is denoted by $\upsilon ( x_{t} | \rho , \textrm{ } \pi , \textrm{ } t )$. This distribution can be computed iteratively using the following formula $\upsilon \left(x_{t + 1}|\rho , \textrm{ } \pi , \textrm{ } t + 1\right) = \int_{\mathcal{X}}^{} P_{\pi} ( x_{t + 1} | x_{t} ) \upsilon ( x_{t} | \rho , \textrm{ } \pi , \textrm{ } t ) d x_{t}$, $\forall t \in \mathbb{N}$,  with the initial condition $\upsilon ( x_{0} | \rho , \textrm{ } \pi , \textrm{ } 0 ) = \rho$, which represents the initial state distribution.

\textit{\textbf{Observations, actions, and rewards}}: Similar to \cite{7}, we implement a three-layer multilayer perceptron (MLP) policy, which processes an observation vector comprising both task-independent and task-relative components:
\begin{equation}
\vec{o} = \left\{ \vec{p}_{\text{off}}, \vec{v}_{\text{cur}}, h_d, q_t, \vec{\omega}_t, \vec{o}_{\text{obs}}, \vec{o}_{\text{task}} \right\},
\end{equation}
where $\vec{p}_{\text{off}}$ denotes the positional offset between the target position and the AUV's current location, $\vec{v}_{\text{cur}}$ represents the water velocity, $h_d$ indicates the water depth, $q_t$ specifies the orientation quaternion, and $\vec{\omega}_t$ represents the measured angular velocities. All positional variables are defined in the AUV's body-fixed coordinate system to ensure goal-oriented control. Additionally, $\vec{o}_{\text{obs}}$ facilitate obstacle avoidance, while $\vec{o}_{\text{task}}$ represents task-specific observations (e.g. positions for other AUVs, for multi-AUV tasks).

For high-level decision-making, the policy generates reference control signals:
\begin{equation}
\vec{a} = [\theta_t, \dot{\psi}_t, n_t],
\end{equation}
where $\theta_t$ denotes the target pitch angle, $\dot{\psi}_t$ represents the target yaw rate, and $n_t$ specifies the target rotational speed of thrusters for velocity control. These reference signals enable direct comparison with observations, providing actionable inputs for S-Surface controllers.

The reward function provides performance feedback for policy optimization. To facilitate subsequent LLM-based adaptation, the study defines a weighted reward structure: 
\begin{equation}
\mathcal{R}_\pi = \boldsymbol{\lambda}^T \boldsymbol{R} = \sum_{i=1}^p \lambda_i R_i,
\end{equation} 
where $\boldsymbol{R} = \{R_1, R_2, \ldots, R_p\}$ represents distinct objectives (e.g., positional accuracy, orientation control, and energy efficiency), and $\boldsymbol{\lambda} = \{\lambda_1, \lambda_2, \ldots, \lambda_p\}$ denotes their corresponding weights.

\textit{\textbf{S-Surface Controller}}: The RL policy generates adaptive reference signals, requiring precise tracking by S-Surface controllers. Each S-Surface controller computes the control signal $u_t$ based on the error $e$ and its derivative $\dot{e}$ between the reference and actual states:  
\begin{equation}
u_t = \frac{2}{1 + \exp(-\zeta_1 e - \zeta_2 \dot{e})} - 1 + \underbrace{\Delta u}_{\text{disturbances}},
\end{equation}
where $\zeta_1$ and $\zeta_2$ are positive constants that serve as surface coefficients. The term $\Delta u$ accounts for environmental disturbances identified by the environment-aware module. The S-Surface's nonlinear exponential component ensures finite-time convergence and provides a smooth control signal.

\subsection{LLM-enhanced Iterative Joint Optimization Module}

For RL-driven control systems to achieve effective performance, both the controller and the reward function must provide explicit performance feedback \cite{22, 36}, as their coupled relationship presents significant tuning challenges. To address this, we propose a joint optimization of the reward function and controller parameters. The optimization objective is formulated as follows:
\begin{equation}
    \underset{\boldsymbol{\lambda},\zeta_1,\zeta_2}{\mathrm{argmax}} \underset{T \rightarrow \infty}{\mathrm{lim}} \mathbb{E}_{\pi} \left[\sum_{t = 0}^{T}\gamma^{t} \Upsilon\left(\boldsymbol{R} \left(\pi , \boldsymbol{\lambda},  \zeta_{1 ,} \zeta_{2}\right)\right)\right],
\end{equation}
where \( \Upsilon:\mathbb{R}^p \to \mathbb{R} \) is a utility function that maps multi-dimensional rewards to a scalar value \cite{47}. While the scalarization process is not fixed and varies with user needs across different scenarios and over time, we maximize \( \Upsilon \) indirectly through performance logs, hard safety constraints, and task prioritization.

Module (B) of the Fig. 2 illustrates the LLM-Enhanced Iterative Joint Optimization Module. Environmental specifications and decomposed user requirements, such as performance metrics and safety constraints, form the context. RL training logs, including performance metrics, guide reward adjustments, while signal tracking performance guides controller adjustments. However, traditional controller tuning metrics, such as settling time and phase margin, struggle to handle RL-generated reference signals characterized by high variability and noise. Therefore, we use visual signal tracking results as inputs. The LLM analyzes tracking performance across critical signal phases, such as steady-state and transients, and diagnoses issues like overshoot, sluggish response, or oscillations. Controller parameters, specifically \( \zeta_1 \) and \( \zeta_2 \) for the S-surface controller, are adjusted based on their physical interpretations.

To mitigate context overload in LLM reasoning, we implement a memory-augmented parameter tracking module using separate visual LLMs. This submodule processes historical parameter-performance correlations, generates comparative summaries, and determines whether it is necessary to terminate optimization if enough optimization or  controller limits are detected.

For efficient joint optimization, a bottleneck-driven synchronization strategy is introduced: the system identifies whether performance limitations stem from reward function misalignment or controller inadequacy, then prioritizes adjustments to reward parameters ($\boldsymbol{\lambda}$), controller parameters ($\zeta_1,\zeta_2$), or both. And finally, the LLM generates formatted output for parameter adjustment. Besides, the reward weights will undergo preliminary tuning based on training feedback under ideal environments (allowing the RL policy to directly adjust the positions of the AUVs without control characteristics), thereby speeding up the adjustment in the controll-constraint scenarios.

\subsection{Environment-Aware and Simulation Module}
To achieve realistic 6-DoF simulation, we utilize the Python Vehicle Simulator \cite{39} based on Fossen’s motion equations \cite{42}, which is capable of simulating real-world hydrodynamic and hydrostatic forces, while providing high-level control input interfaces.

To evaluate AUV disturbance rejection, we simulate marine environments including waves and currents. The fetch-limited JONSWAP (Joint North Sea Wave Project) spectrum is adopted to represent wave energy distribution \cite{40}:
\begin{equation}
S(f) = \frac{\alpha g^2}{(2\pi)^4 f^5} \exp\left(-\frac{5}{4}\left(\frac{f_p}{f}\right)^4\right)\gamma^{\exp\left(-\frac{(f - f_p)^2}{2\sigma^2 f_p^2}\right)},
\end{equation}
where $\alpha$ denotes the energy scale parameter, $f_p$ represents the peak frequency, $\gamma$ is the peak enhancement factor, and $\sigma$ is the peak shape parameter, defined as $\sigma = \sigma_a$ for $f \leq f_p$ and $\sigma = \sigma_b$ for $f > f_p$. The parameter values are listed in Table I. Then, Wave surfaces are generated through linear superposition\cite{41}:
\begin{equation}
\eta(x, y, t) = \sum_{i,j} a_{ij} \cos(\varphi_{ij}),
\end{equation}
\begin{equation}
\varphi_{ij} = k_{ij}x\cos\theta_j + k_{ij}y\sin\theta_j - \omega_i t + \phi_{ij},
\end{equation}
using directional spreading function $D(\theta_j) = \cos^2\theta_j$ and phase offsets $\phi_{ij}$ sampled from a Gaussian process. Component amplitudes derive from $a_{ij} = \sqrt{2S(f_i)D(\theta_j)\Delta f \Delta\theta}$, where $\Delta f$ and $\Delta\theta$ represent frequency/directional resolutions. The dispersion relation $(2\pi f)^2 = gk \tanh(kh)$ determines wave numbers $k_{ij}$. Horizontal wave-induced flows follow Airy theory:
\begin{equation}
\vec{v} = \begin{pmatrix} v_x \\ v_y \end{pmatrix}\! = \!\sum_{i,j} a_{ij} \omega_i \frac{\cosh\left[k_{ij}(z+h)\right]}{\sinh(k_{ij}h)} \cos(\varphi_{ij}) \begin{pmatrix} \cos\theta_j \\ \sin\theta_j \end{pmatrix},
\end{equation}
where $h$ denotes the water depth. Although vertical flow disturbances are currently excluded, wave-induced coupling effects still pose challenges for 6-DoF control due to AUV motion dynamics. To address this, we design an \textbf{Environment-Aware Module}. The AUVs are equipped with horizontal acoustic Doppler current profilers (H-ADCPs) to measure water velocities and active sonar systems for terrain and obstacle detection to avoid collision. Additionally, unmanned surface vehicles (USVs) are utilized to estimate AUV positions via ultra-short baseline (USBL) acoustic positioning and facilitate inter-vehicle communication \cite{43}.

\section{EXPERIMENTS AND ANALYSIS}

In the following, we first describe our simulation setup and then evaluate and analyze the adaptability and performance of our proposed LLM-enhanced RL-based adaptive S-surface controller through comprehensive experiments.

\subsection{Experiment Setup}
We validate the effectiveness of our proposed controller utilizing a REMUS 100 AUV (1.6 m in length, 31.9 kg in weight) with a maximum disturbance-free velocity of 2.3 m/s. The terrain data are derived from the East China Sea region (123°E–124°E, 28°N–29°N), and is post-processed to reduce depth variations, with the deepest water reaching 60m. Additionally, we use the TD3 as our RL algorithm with default settings \cite{37}. Key experiment parameters and configurations are summarized in Table 1.

\begin{table}
  \centering
  \caption{\small Key parameters of the experimental setup.} 
  \label{tab:feature-splitting}

    \begin{tabular}{lc}
      \toprule
      \textbf{Parameters} & \textbf{Values} \\
      \midrule
        JONSWAP parameters  & \multirow{2}{*}{0.01,0.1,3.3,0.07,0.09} \\
        $\alpha$,$f_p$,$\gamma$, $\sigma_a$, $\sigma_b$ & \\
        AUV maximum speed $v_{\text{max}}$, $\omega_{\text{max}}$ & 2.3m/s(4.5kts), 15deg/s(0.26rad/s)\\
        Propeller maximum revolution & 1525rpm \\
        Water density $\rho$ & 1026$\text{kg/m}^{\text{3}}$ \\
        Control frequency & 20Hz \\
        \multirow{2}{*}{LLM model} & GPT-4o (VLLM) \\
        & deepseek-V3 (Textual) \\
        LLM parameters & temperature=0.5, Top P=1 \\
      \bottomrule 
  \end{tabular}
\end{table}

Within this specific setup, we introduce two high-level tasks, whose description is outlined as follows:

• \textbf{3D data collection task}: Employing the proposed controller, a single or multiple AUVs operate together to search and collect data from sensor nodes (SNs) scattered randomly. The main objectives contain conducting adaptive control of AUVs to optimize data collection rates, reducing energy consumption, and enhancing the capability to avoid collisions. We refer further details on this task to \cite{11}.

• \textbf{3D target tracking task}: A single or multiple AUVs are utilized to follow a dynamic underwater target whose position is unpredictable. Other task objectives include avoiding collisions with hazardous terrain and obstacles, maintaining a reasonable water depth, and maintaining communication between AUVs (if applicable). We refer more details to \cite{45}.

\subsection{Experimental Results}

To evaluate the joint optimization of the LLMs, we perform parameter adjustments that the controller parameters are previously set to under-regulation configurations. The results of 3D data collection tasks executed during optimization are illustrated in Fig. 3. For comparative analysis, we also utilize the S-Surface controller to track fixed reference control signals obtained from a target tracking task during optimization, with the comparative tracking performance shown in Fig. 4.

\begin{figure}[!t]
        \centering
        \includegraphics[width=0.981\linewidth]{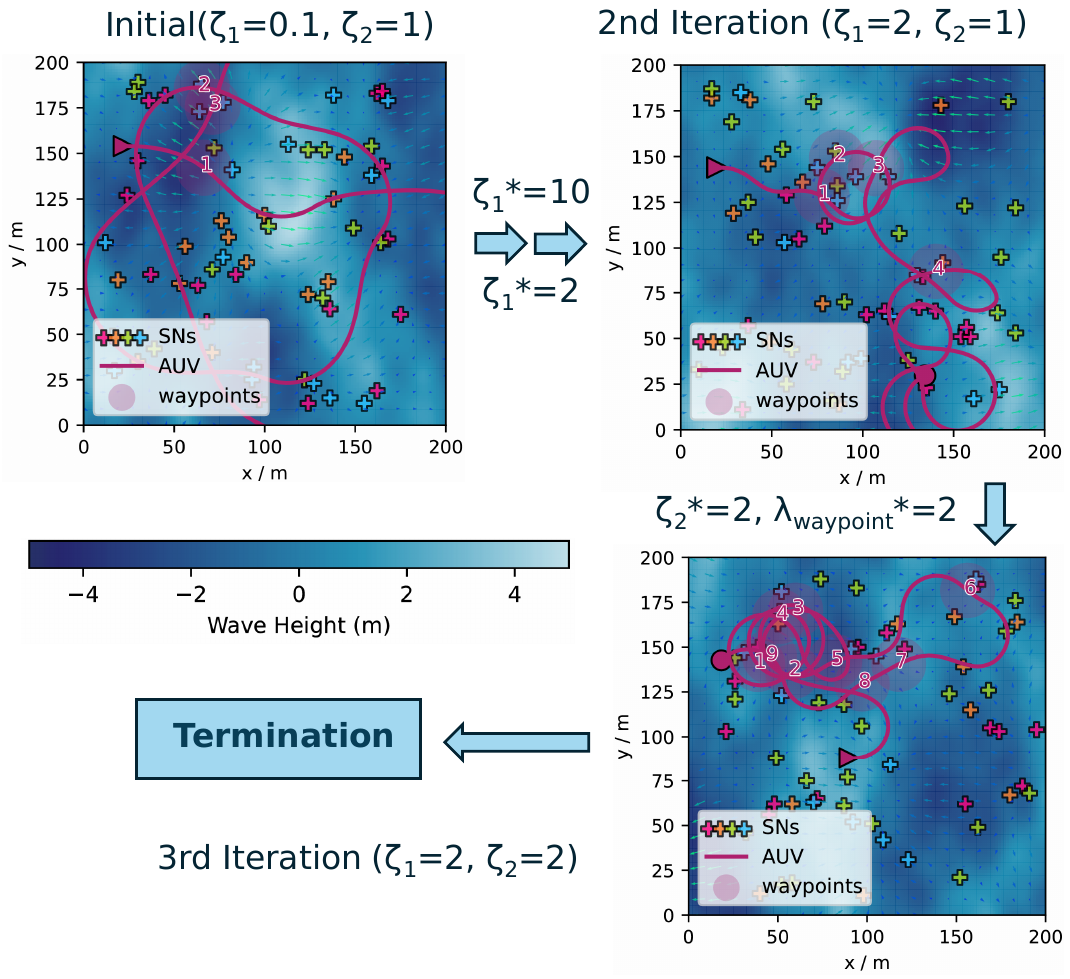}
        \caption{\small Parameters for yaw tracking controller and reward weights, along with 2D projections of AUV trajectories from the 3D data collection tasks during the LLM optimization phase.}  %
        \label{fig_3}
        \end{figure}

\begin{figure}[!t]
        \centering
        \includegraphics[width=0.983\linewidth]{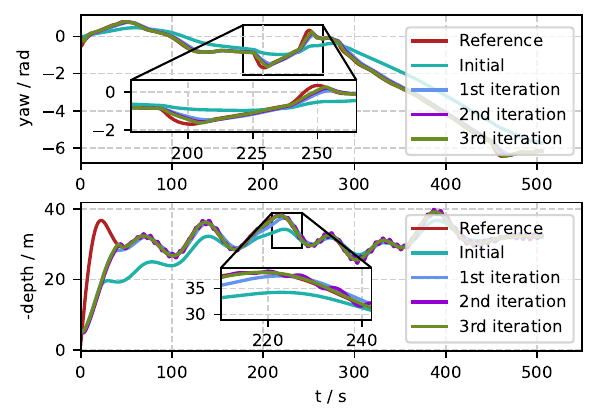}
        \caption{\small Comparative results of the S-surface controller in tracking reference signals taken from a target tracking task during the LLM optimization phase.} 
        \label{fig_4}
        \end{figure}

For yaw control, a low $\zeta_1$ parameter value results in a significantly slow response. Consequently, the LLM substantially increases $\zeta_1$ during the first iteration. In the second iteration, while continuing to increase $\zeta_1$, the adjustment magnitude is reduced due to improved tracking performance. By the third iteration, residual inadequate regulation in yaw control persists when the reference signal shows rapid change, indicating system steering limitations. The LLM responds by increasing $\zeta_2$ before terminating the iteration to enhance stability. Concurrently, it enhances the reward weight for waypoint tracking, as the AUV struggles with flexible and accurate waypoint tracking in practical tasks. For depth control, high-frequency oscillation prompts the LLM to reduce $\zeta_1$ while increasing $\zeta_2$ before termination.

Also, We conduct comparative experiments between the LLM-optimized S-Surface controller and baseline controllers, with their parameters also been optimized by the LLM process mentioned before, including:
\begin{itemize}
    \item \textbf{PID}: Conventional PID controllers for separate yaw and depth control.
    \item Original control implementation from Python Vehicle Simulator (denoted as \textbf{PVS}): The PVS employed a SMC controller with reference model compensation for yaw control and a PI controller for depth control.
\end{itemize}

These controllers are evaluated under two disturbance conditions: the extreme sea condition (\textbf{ES}) with a maximum water velocity of 2 m/s, and the very extreme sea condition (\textbf{VES}) with doubled water velocities (maximum 4 m/s), exceeding the AUV's maximum propulsion capability and requiring advanced compensation strategies.

\begin{figure}[!t]
        \centering
        \subfigure[ES condition]{
        \includegraphics[width=0.48\linewidth]{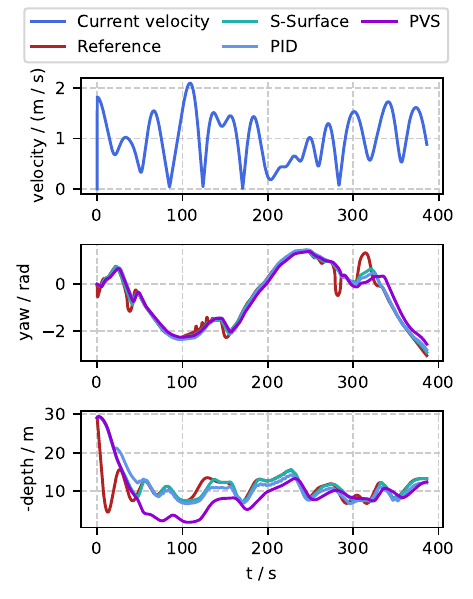}}
        \hspace{-2mm}
        \subfigure[VES condition]{
        \includegraphics[width=0.48\linewidth]{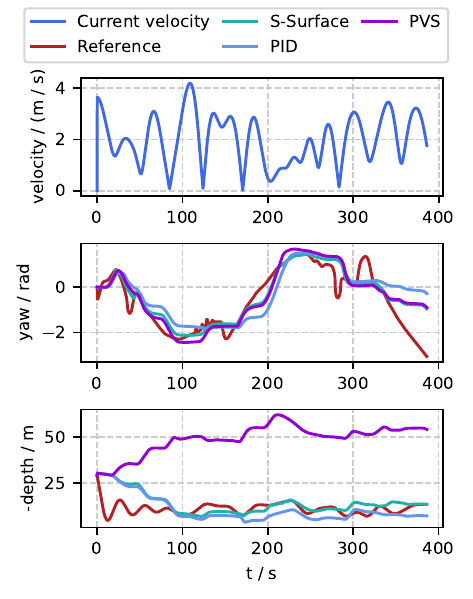}}
        \caption{\small Comparative results of three controllers tracking reference signals taken from a target tracking task under ES and VES conditions.}
        \end{figure}

Comparative results illustrated in Fig. 5 demonstrate the S-Surface controller's superior adaptability under disturbances. Specifically, both the PID and S-Surface controllers achieve stable reference tracking under ES conditions, while the PVS exhibits delayed yaw control response due to the inherent phase lag of its SMC controller with reference model architecture, along with significant depth overshoot from its basic PI controller. When transitioning to VES conditions, the controllers exhibit progressive performance deterioration, with the PID controller showing worse flexibility and stability compared to the S-Surface controller. Additionally, the PVS suffers a complete loss of depth regulation capability.

Then, we utilize the controllers above to conduct 3D data collection task to evaluate the task-specific performance. We introduce three metrics: the total number of served sensor nodes (\textbf{SSN}, quantifying yaw control capability), energy consumption (\textbf{EC}, measuring actuation efficiency, calculated using equations from \cite{44}), and danger time (\textbf{DT}, representing the cumulative duration of unsafe seafloor proximity below 10 m, quantifying depth control capability). An idealized control setting (\textbf{Ideal}) is additionally introduced, which removes hydrodynamic limitations and the RL policy can directly change the AUVs' positions. The results are presented in Table II. Under the ES condition, the S-Surface controller achieves performance close to the ideal setting, while the PVS exhibits significantly longer danger time due to poor depth control. Under the VES condition, the PID controller exhibits significantly greater performance degradation compared to the S-Surface controller, while the PVS experiences serious control failure.

\begin{table}
  \centering
  \caption{\small Performance metrics of different control methods evaluated during the data collection task under ES and VES conditions.} 

    \begin{tabular}{ccccc}
      \toprule
      \multicolumn{2}{c}{Metrics} & \textbf{SSN} $\uparrow$ & \textbf{EC} (W) $\downarrow$ & \textbf{DT} (s) $\downarrow$\\
      \midrule
      \multicolumn{2}{c}{\textbf{Ideal}} & 15.7 $\pm$ 6.4 & 163.8 $\pm$ 32.0 & 0.0 $\pm$ 0.0 \\
      \midrule
    \multirow{2}{*}{\textbf{S-Surface}} & ES & 14.0 $\pm$ 8.7 & 202.9 $\pm$ 38.5 & 0.0 $\pm$ 0.0 \\
    & VES & 10.7 $\pm$ 7.0 & 227.2 $\pm$ 44.5 &34.4 $\pm$ 20.7 \\
    \midrule
    \multirow{2}{*}{\textbf{PID}} & ES & 13.8 $\pm$ 9.0 & 194.8 $\pm$ 41.9 &0.0 $\pm$ 0.0 \\
    & VES & 9.2 $\pm$ 6.1 & 231.3 $\pm$ 46.8 &53.7 $\pm$ 23.3 \\
    \midrule
    \multirow{2}{*}{\textbf{PVS}} & ES & 12.3 $\pm$ 8.4 & 205.1 $\pm$ 35.2 &203.0 $\pm$ 98.8 \\
    & VES & 6.5 $\pm$ 5.4 & 247.1 $\pm$ 56.3 &517.8 $\pm$ 157.9 \\
      \bottomrule 
  \end{tabular}
\end{table}

\begin{figure}[!t]
        \centering
        \subfigure[3D Data collection task]{
        \includegraphics[width=0.51\linewidth]{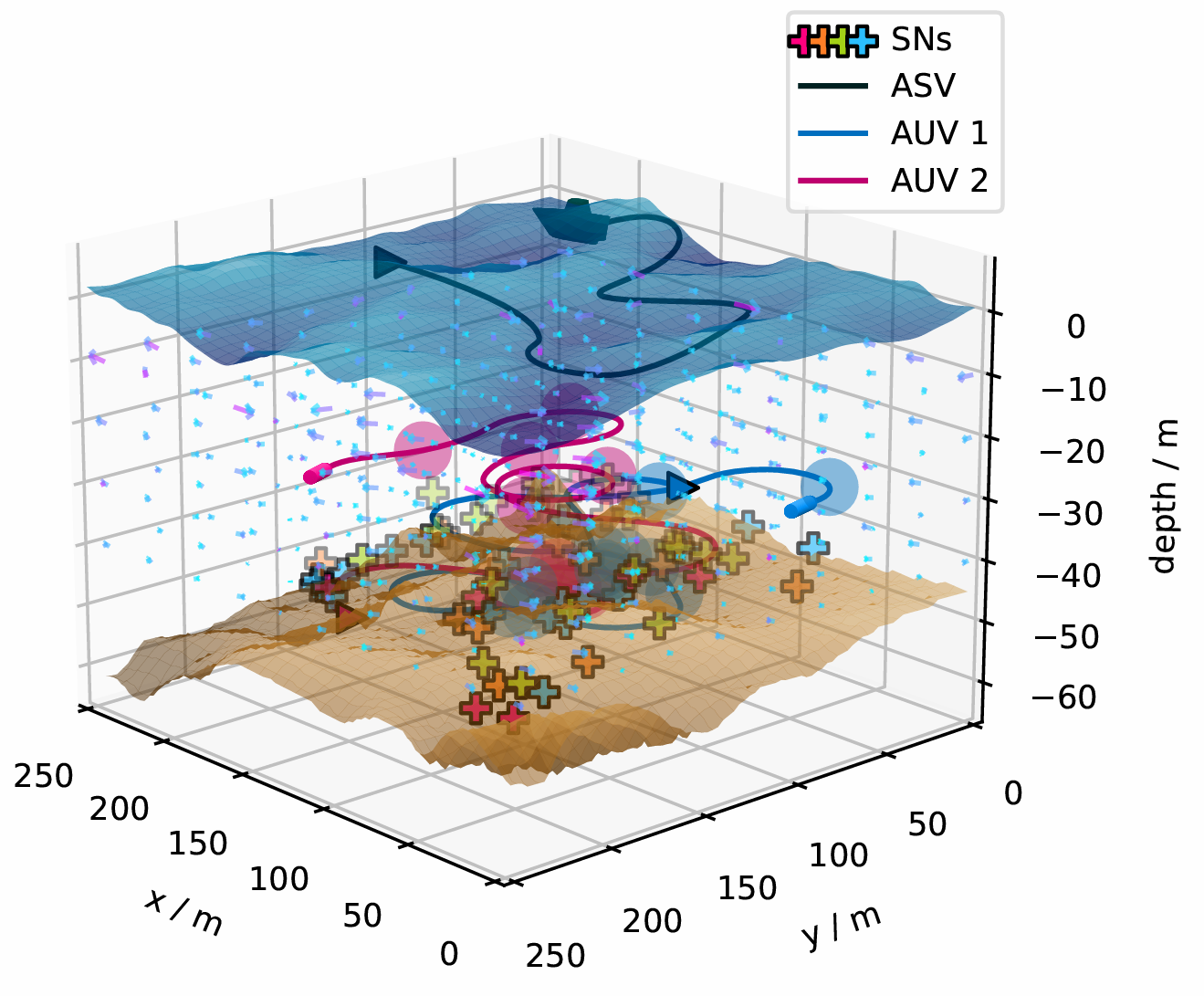}}
        \hspace{-4mm}
        \subfigure[3D target tracking task]{
        \includegraphics[width=0.475\linewidth]{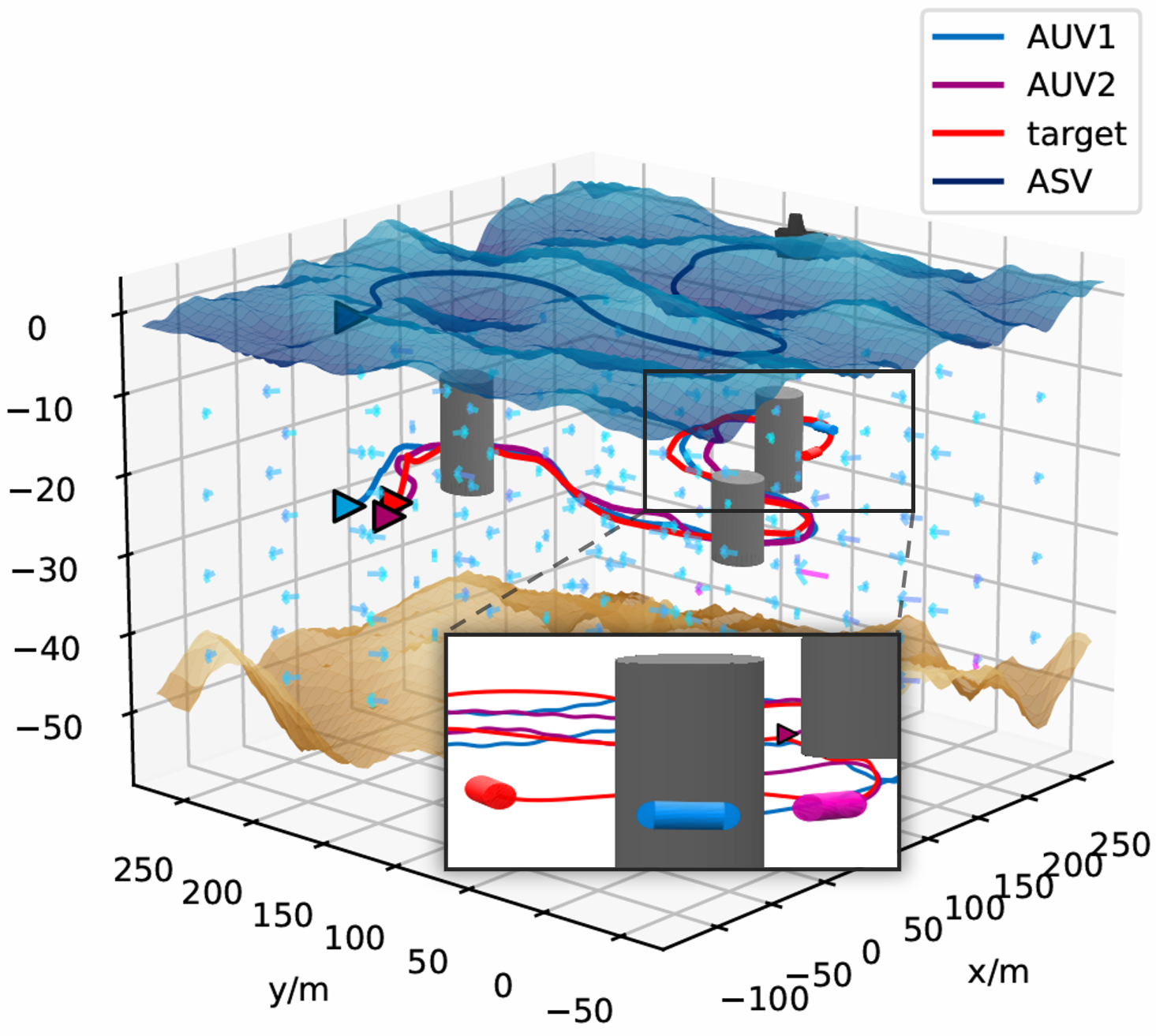}}
        \caption{\small 3D visualizations of multiple AUVs performing data collection and target tracking tasks using the proposed RL-based S-surface controller.}
        \end{figure}

Finally, Fig. 6 visualizes the 3D data collection and target tracking tasks performed by two AUVs utilizing S-Surface control. In the former case, the AUV must judiciously control its direction to efficiently serve sensor nodes due to its restricted tuning capability, while the latter requires more real-time control capabilities. Thanks to the powerful optimization capability of RL and the flexible execution of the controller, the AUVs can plan optimal routes as much as possible, achieving performance close to ideal control conditions. In the latter case, the AUVs also demonstrate high maneuverability in response to target turns.

\section{CONCLUSIONS}
In this study, we develop an LLM-enhanced RL-based adaptive S-surface controller for AUVs under extreme sea conditions. This controller utilizes LLMs to iteratively optimize controller parameters and reward functions, while leveraging RL to train the AUV to acquire an expert-level control strategy. The strategy autonomously generates control commands for S-surface controllers in high-level tasks, which further convert them into low-level control signals. Comprehensive simulation experiments on representative high-level tasks demonstrate the superior performance and adaptability of the proposed controller, which outperforms PID and SMC controllers under extreme sea conditions. Future work will focus on implementing the proposed controller on AUVs and conducting field experiments to realize the sim2real process, aiming to minimize the gap between simulation and reality.

\section{ACKNOWLEDGEMENT}
The authors gratefully acknowledge the anonymous reviewers for their constructive comments. We also extend our sincere thanks to Dr. Xiangwang Hou, Prof. Yong Ren, Prof. Daoyi Chen, and Prof. Juntian Qu from Tsinghua University for their insightful discussions and guidance. Additionally, we appreciate the encouragement and recognition from Prof. Nare Karapetyan at the Woods Hole Oceanographic Institution, Prof. Xiaofan Li at the University of Hong Kong, and Prof. Xiaomin Lin at the University of South Florida.





\bibliographystyle{IEEEtran}
\bibliography{IEEEexample}

\addtolength{\textheight}{-12cm}   

\end{document}